\documentclass[lettersize,journal]{IEEEtran}
\usepackage{graphicx}
\usepackage{amsmath}
\usepackage{makecell}
\usepackage{filecontents}
\usepackage{times}
\usepackage{epsfig}
\usepackage{graphicx}
\usepackage{amsmath}
\usepackage{amssymb}
\usepackage{threeparttable}  
\usepackage{float}
\usepackage{epsfig}
\usepackage{graphicx}
\usepackage{amsmath}
\usepackage{amssymb}
\usepackage{balance}
\usepackage{booktabs}
\usepackage{color}
\usepackage{enumitem}
\usepackage{mathtools}
\usepackage{makecell}
\usepackage{multirow}
\usepackage{pifont}
\usepackage{subfigure}
\usepackage{stackengine}

\usepackage[table]{xcolor}
\usepackage{soul}
\usepackage{array}
\usepackage{subfigure}
\usepackage{subcaption}
\usepackage{algorithm}
\usepackage{algorithmic}
\usepackage[hidelinks]{hyperref}
\usepackage{tabularx}
\usepackage{makecell}
\usepackage{pythonhighlight}
\usepackage{xcolor}
\usepackage{balance}

\newcolumntype{C}{>{\centering\arraybackslash}X}

\renewcommand{\arraystretch}{1.2}

\begin{document}

\title{LC-LLM: Explainable Lane-Change Intention and Trajectory Predictions with Large Language Models} 

\author{Mingxing Peng, Xusen Guo, Xianda Chen, Meixin Zhu*, and Kehua Chen
\thanks{Manuscript received XX June, 2024.}
\thanks{Corresponding author is Meixin Zhu (E-mail: meixin@ust.hk).}
\thanks{Mingxing Peng, Xusen Guo and Xianda Chen are with the Systems Hub, The Hong Kong University of Science and Technology (Guangzhou). Meixin Zhu is with the Systems Hub, The Hong Kong University of Science and Technology (Guangzhou) and Guangdong Provincial Key Lab of Integrated Communication, Sensing and Computation for Ubiquitous Internet of Thing. Kehua Chen is with the Division of Emerging Interdisciplinary Areas (EMIA), Academy of Interdisciplinary Studies, The Hong Kong University of Science and Technology, Hong Kong, China.}}

\markboth{IEEE Transactions on Intelligent Vehicles,~Vol.~xx, No.~x, Xxx~2024}%
{Shell \MakeLowercase{\textit{et al.}}: A Sample Article Using IEEEtran.cls for IEEE Journals}


\maketitle
\begin{abstract}
To ensure safe driving in dynamic environments, autonomous vehicles should possess the capability to accurately predict lane change intentions of surrounding vehicles in advance and forecast their future trajectories. Existing motion prediction approaches have ample room for improvement, particularly in terms of long-term prediction accuracy and interpretability. In this paper, we address these challenges by proposing LC-LLM, an explainable lane change prediction model that leverages the strong reasoning capabilities and self-explanation abilities of Large Language Models (LLMs). Essentially, we reformulate the lane change prediction task as a language modeling problem, processing heterogeneous driving scenario information as natural language prompts for LLMs and employing supervised fine-tuning to tailor LLMs specifically for lane change prediction task. 
Additionally, we finetune the Chain-of-Thought (CoT) reasoning to improve prediction transparency and reliability, and include explanatory requirements in the prompts during inference stage. Therefore, our LC-LLM model not only predicts lane change intentions and trajectories but also provides CoT reasoning and explanations for its predictions, enhancing its interpretability.
Extensive experiments based on the large-scale highD dataset demonstrate the superior performance and interpretability of our LC-LLM in lane change prediction task. To the best of our knowledge, this is the first attempt to utilize LLMs for predicting lane change behavior. Our study shows that LLMs can effectively encode comprehensive interaction information for driving behavior understanding.
\end{abstract}


\section{Introduction}
\IEEEPARstart{L}{ane} change prediction is a critical task for autonomous driving systems to ensure safe and efficient navigation on the road. Lane changing behavior, which refers to the action of a vehicle moving from one lane to another, is one of the most complex driving behaviors due to intricate interactions with other vehicles. Accurately forecasting the underlying lane change intentions and future trajectories of surrounding vehicles in advance is essential for autonomous vehicles to make informed decisions and take appropriate actions. 

In recent years, considerable advancements have been made in lane change predictions. Various approaches have been explored to enhance performance in detecting lane change maneuvers \cite{he2019probabilistic, mozaffari2022early, xin2018intention, gao2023dual}. However, at the early stage of a lane change maneuver when lateral displacement is minimal, it is essential to rely on the interaction information between the target vehicle and its surrounding vehicles for accurate predictions. Therefore, the ability to effectively model interaction information among vehicles becomes paramount. Previous methods exhibit limited capability in understanding interactive information, which results in weak long-term prediction performance \cite{he2019probabilistic, xin2018intention}. 
Additionally, lane change trajectory prediction has witnessed notable progress with the adoption of deep learning techniques, such as Graph Neural Networks (GNN) and Transformers, leading to competitive results \cite{gao2023dual, shi2022motion, seff2023motionlm}. Despite these successes, these deep learning-based approaches frequently suffer from a lack of interpretability, as they generate predictions about future behaviors without offering substantial explanations for their results. In summary, enhancing the capability to model interactions among vehicles and augmenting the interpretability of prediction results are key challenges in lane change prediction for autonomous driving.

In recent years, advancements in Large Language Models (LLMs) have provided new opportunities for addressing challenges associated with lane change prediction. Noteworthy progress in LLMs \cite{achiam2023gpt, ouyang2022training, touvron2023llama} has showcased their robust information comprehension skills and powerful common sense reasoning abilities. 
However, current researches on lane change prediction have not yet incorporated LLMs. Given the LLMs' profound understanding of complex driving scenarios and their extensive knowledge base, including common driving knowledge, it is rational to utilize LLMs in modeling interactions between vehicles for the enhancement of future driving behavior predictions. In addition, LLMs have demonstrated the capacity to leverage their vast knowledge base to generate explanations for their predictions \cite{rajani2019explain, huang2023can}. Inspired by these studies, there is potential to leverage LLMs for the explanation of predicted lane change intentions and future trajectories, thereby augmenting the interpretability of predictions in autonomous driving.

In this paper, we propose a novel approach that leverages LLMs' powerful reasoning and self-explanation capabilities to address lane change prediction challenges mentioned above. In essence, we reformulate the task of predicting lane change intentions and trajectories as a language modeling problem. To this end, we integrate heterogeneous inputs, including target vehicle state, surrounding vehicle state, and map information, converting them into structured prompts in natural language to input into the LLM. Then, we employ a supervised fine-tuning technique to tailor the LLM specifically for the lane change prediction task. 
Additionally, in fine-tuning stage, we perform Chain-of-Thought (CoT) reasoning to enhance transparency and reliability in lane change predictions. Finally, we integrate explanatory requirements into the prompts in the inference stage, thus enabling our fine-tuned model to generate explanations for the lane change prediction results.
Benefiting from the powerful capabilities of LLM, our fine-tuned model demonstrates better performance and interpretability in lane change prediction.  

The main contributions of this paper include: 
\begin{itemize}
    \item We propose LC-LLM, the first Large Language Model for lane change prediction. It leverages the powerful capabilities of LLMs to understand complex interactive scenarios, enhancing the performance of lane change prediction. 
    \item Our LC-LLM achieves explainable predictions. It not only predicts lane change intentions and trajectories but also generates explanations for the prediction results. 
    Additionally, we evaluate the accuracy of CoT reasoning to quantitatively evaluate the interpretability of our LC-LLM.
    \item Extensive experiments on the highD dataset demonstrate that our LC-LLM outperforms all baseline models, achieving a 17.7\% improvement in lane change intention prediction, a 64.4\% improvement in lateral trajectory prediction, and a 66.1\% improvement in longitudinal trajectory prediction, respectively. 
\end{itemize}
The remainder of the paper is organized as follows. Section \uppercase\expandafter{\romannumeral2} reviews previous research related to our work. Section \uppercase\expandafter{\romannumeral3} presents the LLM-based lane change intentions and trajectory prediction approach. Section \uppercase\expandafter{\romannumeral4} provides experimental results demonstrating the performance of the proposed approach. Finally, the conclusions are drawn in Section \uppercase\expandafter{\romannumeral5}.

\section{RELATED WORK}
The literature reviews \cite{huang2022survey, mozaffari2020deep, chen2022milestones} discuss various methods for predicting the future states of vehicles for autonomous driving systems. These prediction studies can be classified into the following two categories based on the output type: Intention prediction and trajectory prediction. Concurrently, LLMs have shown rapid progress in recent developments. Comprehensive overviews of LLMs applications in autonomous driving are discussed in \cite{yang2023survey}. 
\subsection{Intention Prediction}
Lane change intention prediction is a critical component in the domain of autonomous driving and advanced driver assistance systems (ADAS). Research in this field focuses on accurately predicting vehicles' intention to change lanes before the maneuver is executed, thereby enhancing road safety and traffic flow. Various approaches have been explored, including machine learning techniques, probabilistic modeling, and deep learning. Mandalia et al. \cite{mandalia2005using} first applied the support vector machine (SVM) to identify lane change maneuvers by incorporating features such as acceleration, steering angle, distance, and so on. Lyu et al. \cite{lyu2020vehicle} predict lane change intentions based on the support vector machine-recursive feature elimination (SVM-RFE) model. He et al. \cite{he2019probabilistic} designed a Dynamic Bayesian Network (DBN) with the purpose of distinguishing between vehicle-following and lane-change maneuvers. Further advancements were made by Hong et al. \cite{hong2019rules} and Mozaffari et al. \cite{mozaffari2022early}, who utilized a convolutional neural network (CNN) as feature extractors to fuse complex driving environment context and effectively learn the driving behavior. Zyner et al. \cite{zyner2017long, zyner2018recurrent} employed Long Short-Term Memory Networks (LSTM) and Recurrent Neural Networks (RNN), respectively, as a sequence classifier for predicting vehicles' driving intentions. Xin et al. \cite{xin2018intention} implemented a dual-block LSTM architecture, where the first LSTM block processes sequential trajectory data to recognize driver intentions as an intermediate indicator. Izquierdo et al. \cite{izquierdo2019experimental} utilized a hybrid CNN-LSTM model to capture both local and global contextual features, as well as temporal information, to forecast vehicle lane change intentions. Gao et al. \cite{gao2023dual} proposed a dual Transformer model, which includes a lane change intention prediction model and a trajectory prediction model. Although these approaches have been proven to have good performance in detecting lane change maneuvers that have already started, the capability of predicting driving intention in advance needs to be improved. 

\begin{figure*}
\centerline{\includegraphics[width=\linewidth, trim= 40 90 2 50, clip]{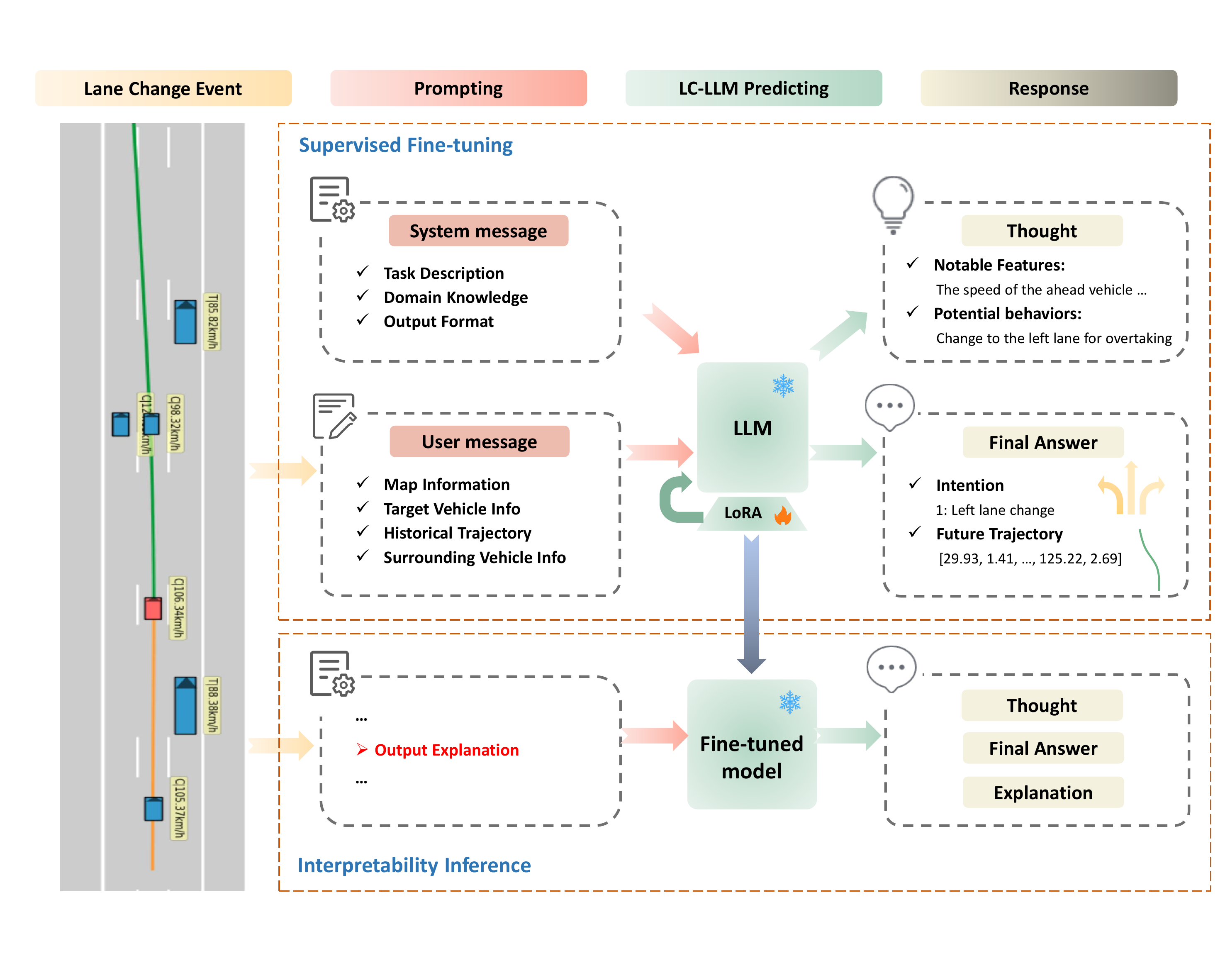}}
\caption{
The pipeline of our LC-LLM. The green trajectory in the Lane Change Event is the future trajectory of the target vehicle, which the model aims to predict. Observations are described using natural language prompts, which are input into the LLM. Supervised fine-tuning methods are then applied to fine-tune the LLM for accurate prediction of lane change intentions and trajectories, as well as the CoT reasoning. During the inference phase, prompts are designed to include explanatory requirements. As a result, the fine-tuned model is capable of predicting the target vehicle's lane change intentions and future trajectory in the current frame, while simultaneously providing the thought reasoning and explanations for its predictions, thus improving the interpretability of the model's outputs.
}
\label{fig:pipeline}
\end{figure*}

\subsection{Trajectory Prediction}
Research on trajectory prediction for autonomous driving is a critical area of study, aiming to enhance the safety and efficiency of self-driving vehicles by forecasting the future trajectory of surrounding entities such as cars, cyclists, and pedestrians. Early approaches used bird's-eye view images as the input and applied a CNN framework to process the rasterized scene for trajectory prediction \cite{chai2019multipath, cui2019multimodal}. Furthermore, several studies employed LSTM networks, utilizing one LSTM as an encoder to capture features from historical trajectories and another LSTM as a decoder to predict future trajectories \cite{deo2018multi, altche2017lstm, xin2018intention}. More recent works represented scenes with vectorized data such as points and polylines, and processed them with GNNs \cite{liang2020learning, gao2020vectornet} or Transformers \cite{shi2022motion, gao2023dual, zhu2022transfollower} to effectively model the interactions between traffic participants and environment. In \cite{gao2020vectornet}, Gao et al. proposed VectorNet which used GNNs to extract features from vectorized high-definition (HD) maps, thereby avoiding lossy rendering and computationally intensive CNN encoding. In \cite{shi2022motion}, Shi et al. proposed the Motion transformer framework, which models motion prediction as the joint optimization of global intention localization and local movement refinement. A recent study \cite{seff2023motionlm} represented continuous trajectories as sequences of discrete motion tokens, framing multi-agent motion prediction as a language modeling task. While these deep learning-based approaches achieve competitive results, their predictions often lack interpretability, which is not conducive to the development of safer and more transparent autonomous driving systems.

\subsection{LLMs for Autonomous Driving}
Recent advancements in LLMs have demonstrated remarkable capabilities in extensive knowledge storage, logical reasoning, and question-answering. Considering these competencies, it is a natural extension to contemplate the application of LLMs to enhance the field of autonomous driving \cite{yang2023survey}. Wu et al. \cite{wu2023language} integrated LLMs with 3D detection and tracking tasks by using language prompts as semantic cues. Mao et al. \cite{mao2023gpt} proposed a prompting-reasoning-finetuning strategy, enabling LLM to generate driving trajectories and showcased the detailed numerical reasoning abilities of GPT-3.5 in motion planning. Chen et al. \cite{chen2023driving} designed an object-level multimodal LLM architecture that combines vectorized numeric modalities with a pre-trained LLM. Additionally, Xu et al. \cite{xu2023drivegpt4} presented an interpretable autonomous driving system employing LLMs and developed a visual instruction tuning dataset for interpretable autonomous driving. While LLMs have been extensively explored in the field of autonomous driving perception and planning, their application in the domains of intention prediction and trajectory prediction remains unexplored.

\section{Methodology}
In this section, we introduce our LC-LLM, a model based on LLM designed for predicting lane change intentions and trajectory in autonomous driving systems. The whole pipeline of our LC-LLM is depicted clearly in Fig.~\ref{fig:pipeline}. We reconceptualize the task of predicting intentions and trajectory as a language modeling problem. To this end, we articulate observations using natural language as prompts for input into the LLM and leverage supervised fine-tuning techniques to tailor the LLM to this specific task. During the inference stage, we incorporate explanatory requirements into the prompt. As a result, our fine-tuned model, LC-LLM, not only forecasts the lane change intentions and future trajectories but also provides CoT reasoning and explanations for the predictions, thereby enhancing their interpretability.

\begin{figure}[ht]
    \centering
    \includegraphics [width=\linewidth, trim= 270 450 275 390, clip] {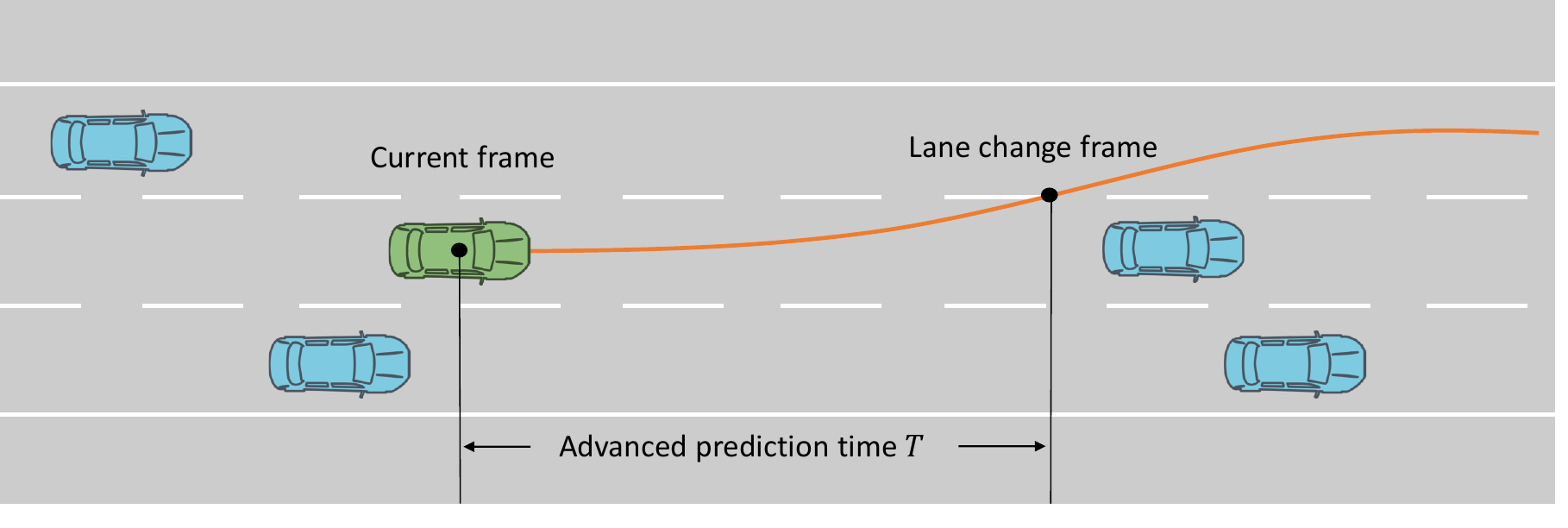}
    \caption{A lane change scenario. The target vehicle is depicted in green and the surrounding vehicles are depicted in blue. The orange line illustrates the trajectory of the target vehicle in the future $t$ timesteps. The advanced prediction time $\textit{T}$ denotes the temporal interval between the lane change frame and the current frame.}
    \label{fig:advanced prediction time}
\end{figure}

\begin{figure*}
\centering
\includegraphics[width=\linewidth, trim= 242 250 230 250, clip]{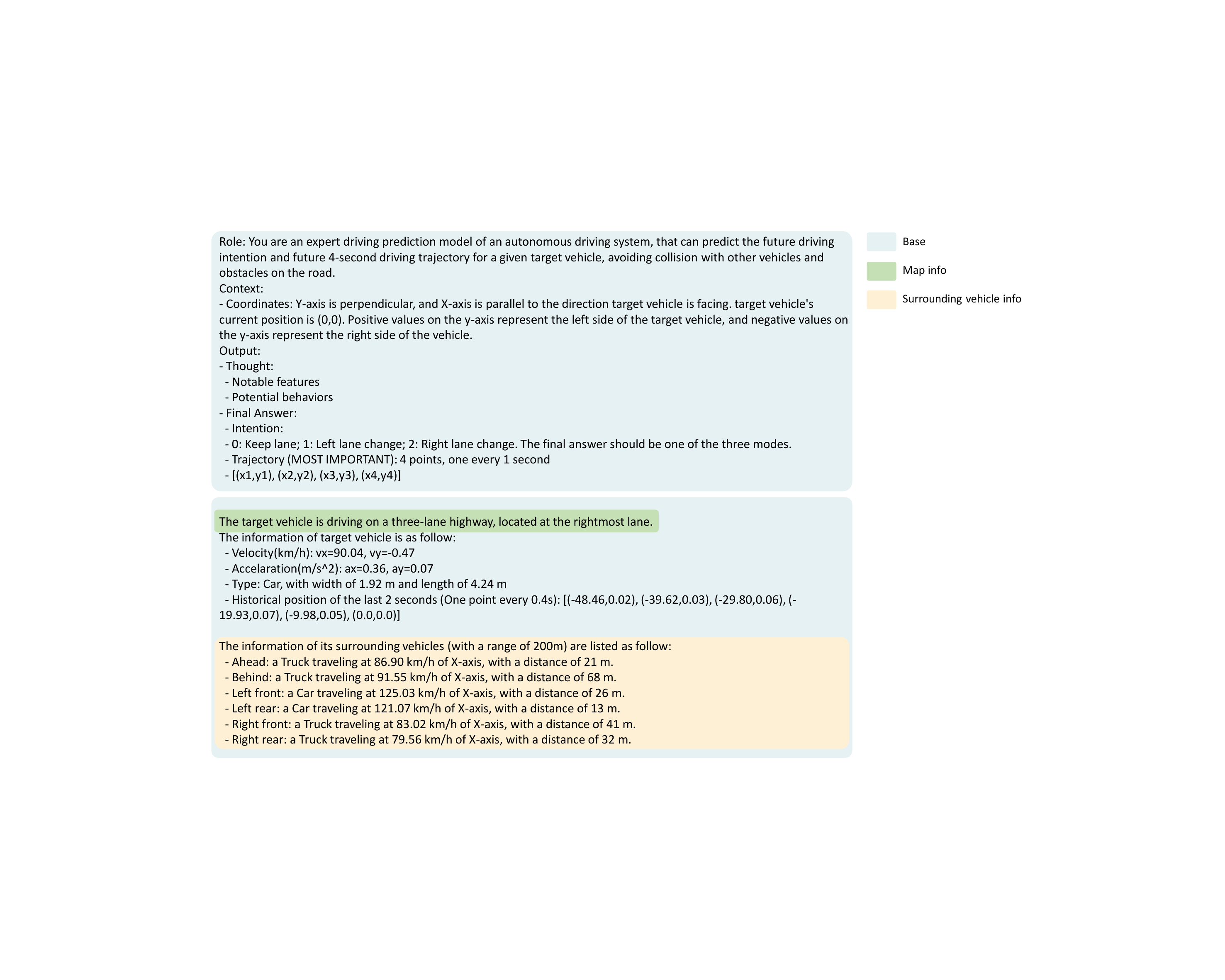}
\caption{An example of our input prompts in the fine-tuning stage. The input prompts comprise a system message displayed in the upper text block and a user message presented in the lower text block. The user message mainly contains map information, the target vehicle's state, the spatial relationships between the target vehicle and its surrounding vehicles.}
\label{fig:prompt}
\end{figure*}

\subsection{Problem Formulation}
In this paper, we aim to develop a predictive model based on LLM that is designed to simultaneously predict the lane change intention and the trajectory of the target vehicle. A representative lane change scenario is depicted in Fig.~\ref{fig:advanced prediction time}, where the target vehicle is illustrated in green and the surrounding vehicles in blue. The objective is to forecast the position of the target vehicle's trajectory at $\textit{t}$ future timesteps, as well as to determine its lane change intentions within the same temporal window, as indicated by the orange line in Fig.~\ref{fig:advanced prediction time}.

We define the state of the target vehicle as $\textbf{\textit{tv}}$, the states of the surrounding vehicles as $\textbf{\textit{sv}}$, and the map information as $\textbf{\textit{m}}$, which serve as the inputs to our model. 
The outputs are the CoT reasoning $\mathcal{C}$, the lane change intention $\mathcal{I}$ and the future trajectory $\mathcal{T}$. The process of the model $\mathcal{F}$ can be formulated as follows:
 \begin{equation}
       \mathcal{C}, \mathcal{I},  \mathcal{T} = \mathcal{F}(\textbf{\textit{tv}}, \textbf{\textit{sv}}, \textbf{\textit{m}})
       \label{definition}
\end{equation}
The $\mathcal{I}\in\{0, 1, 2\}$, where 0 denotes ``keep lane", 1 denotes ``left lane change", and 2 denotes ``right lane change". The future trajectory of prediction horizon $t$ can be 
denoted by $\mathcal{T} = \{(\textit{x}_1, \textit{y}_1), (\textit{x}_2, \textit{y}_2),...,(\textit{x}_t, \textit{y}_t)\}$. And the corresponding CoT reasoning $\mathcal{C}$ is composed of notable features and potential behavious. The target vehicle state $\textbf{\textit{tv}}$ consists of a historical trajectory of this vehicle and its current velocity, type and so on. The surrounding vehicle states $\textbf{\textit{m}}$ contain each vehicle's current velocity, type, and their spatial relationship relative to the target vehicle. The map information $\textbf{\textit{m}}$ includes details such as lane identifiers and lane markings.

We reframe the task of predicting intentions and trajectory as a language modeling problem. By utilizing natural language to describe both the input data and output results, we can represent them as a sequence of tokens. The input data which describe the current frame driving scenario can be denoted as a sequence of tokens  $\textit{T}_{s}$:
\begin{equation}
    \textit{T}_{s} = K(\textbf{\textit{tv}}, \textbf{\textit{sv}}, \textbf{\textit{m}})
\end{equation}
The prediction result which describes the lane change intentions and trajectory can also be represented as a sequence of tokens $\{\textit{T}_1, \textit{T}_2,...,\textit{T}_n\}$:
\begin{equation}
    \{\textit{T}_1, \textit{T}_2,...,\textit{T}_n\} = K(\mathcal{C}, \mathcal{I}, \mathcal{T})
\end{equation}
where the $K$ is a language tokenizer used to transform input data and output results into tokens. The $\textit{T}_i$ represents the $i$-th token in the sequence. By adopting this language-based representation, we can redefine the prediction problem as a language modeling problem, with the loss function resembling that of language modeling \cite{qiu2020pre}:
\begin{equation}
    \mathcal{L} = - \sum_{i=1}^{n} \log \textit{P}(T_{i}^{*} | T_{<i}, T_{s})
    \label{equation:loss}
\end{equation}
where $\textit{T}_{i}^{*}$ denotes the ground truth of the next token given its historical tokens, and $n$ represents the tokens number of ground truth in one sample. By maximizing the conditional probability $\textit{P}$ of the token $\textit{T}_{i}^{*}$, our model can effectively predict lane change intentions and future trajectory.

\subsection{Prompting}
Generally, LLMs receive inputs in the form of natural language rather than unprocessed vectorized data. Consequently, the formulation of effective prompts that describe the current observations in natural language is critical. Several studies have sought to harness the deep reasoning capabilities of the LLMs through clever prompt design \cite{mao2023gpt, cui2023receive}. Building on these prior efforts, we have crafted prompts that are clearer, more intelligent, and better structured. Fig.~\ref{fig:prompt} provides an example of our input prompts. 

As illustrated in Fig.~\ref{fig:prompt}, the input prompts consist of a system message displayed in the upper text block and a user message presented in the lower text block. The system message maintains consistency across diverse driving scenarios. It delineates the designated role of the LLM, provides details of the coordinate system, and outlines the information and format for the LLM's output. In our paper, the assigned role for the LLM is that of a predictive model integrated within an autonomous driving system. The coordinate system in each driving scenario is the vehicle coordinate system, which is centered on the current position of the target vehicle. The expected output includes predictions of lane change intentions and trajectory points over a future time horizon of four seconds, as well as includes thought reasoning which consists of notable features and potential behavior.

The user message provides a description of the observations specific to the current frame, thus varying with each driving scenario. It includes information about the map, the state of the target vehicle, the spatial relationships between the target vehicle and its surrounding vehicles. Map information in the prompts primarily denotes the number of lanes in the scenario and indicates whether the target vehicle is located in the leftmost, middle, or rightmost lane. Prompts related to the target vehicle's state are generated by detailing the target vehicle's historical trajectory over the past two seconds, its current velocity, and its vehicle type. Given that most vehicles do not have a large lateral displacement 4 seconds before changing lanes, the model mainly relies on the interaction information between the target vehicle and surrounding vehicles to predict lane change intentions in advance. Therefore, comprehending the information pertaining to surrounding vehicles is crucial for accurately predicting the lane change intentions of the target vehicle. We denote the information of the nearest vehicles in eight directions surrounding the target vehicle as surrounding vehicle prompts. These directions include the ahead, left front, right front, left side, right side, rear, left rear, and right rear. The surrounding vehicle information in each direction encompasses details such as vehicle type, current speed, and relative distance from the target vehicle. This comprehensive surrounding vehicle information serves as a critical prompt for the predictive model, enabling a thorough analysis of the contextual interactions influencing the target vehicle's lane change intentions.

\begin{figure*}
\centering
\includegraphics[width=\linewidth, trim= 152 600 365 320, clip]{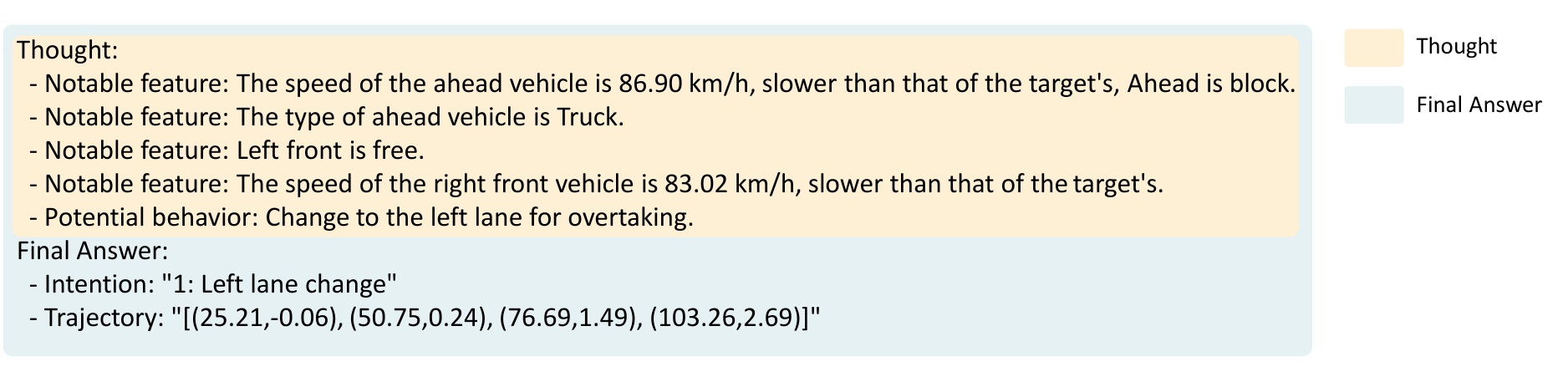}
\caption{An example of the output of our LC-LLM. The content in the yellow text block is the Chain-of-Thought reasoning.}
\label{fig:output}
\end{figure*}

\subsection{Reasoning}
Recently, CoT reasoning has demonstrated remarkable capabilities in carrying out more intricate reasoning tasks \cite{mao2023gpt, wei2022chain, kojima2022large, zhou2022least}. Drawing inspiration from GPT-Driver \cite{mao2023gpt}, which utilizes Chain-of-Thought reasoning strategy in motion planning tasks to enhance transparency throughout the planning procedure, we employ CoT reasoning in lane change prediction task to improve reliability of predictions and explanations. Unlike GPT-Driver, which only labels thoughts based on whether the ego vehicle and surrounding vehicles will collide in the future, we also consider domain knowledge in driving, traffic rules, and traditional lane change model rules \cite{gipps1986model, hidas2002modelling} for labeling CoT reasoning. In this way, our LC-LLM model can learn this knowledge and rules to provide more reliable and accurate predictions and explanations. The details of CoT reasoning are shown in Fig.~\ref{fig:output}.

Our CoT reasoning consists of notable features and potential behaviors. The annotation of CoT reasonings are as follows:
\subsubsection{Labeling of Notable feature}
Firstly, we labeled notable features include significant lateral movement when lateral velocity exceeds $1.5 km/h$, and high longitudinal acceleration when it surpasses $0.4 m/s^{2}$. Secondly, the relative speeds of surrounding vehicles compared to the target vehicle in the directions ahead, left front, and right front are considered notable features. Vehicles in these positions with a speed higher than the target vehicle are labeled as free. Conversely, vehicles with a speed lower than the target vehicle are labeled as blocked, indicating potential congestion or the need for the target vehicle to overtake. At last, the presence of a truck ahead within 100 meters is also notable due to its impact on traffic flow and overtaking maneuvers \cite{gipps1986model}. Additionally, in right lane change scenarios, the characteristic of the target vehicle being a truck is also labeled as notable due to the traffic rule that trucks typically travel in the right lane.

\subsubsection{Labeling of Potential behavior}
Potential behaviors are classified into eight categories: ``Change to the left lane for overtaking" occurs when encountering a slower vehicle ahead and occupying the rightmost or middle lane; ``Change left to the fast lane" corresponds to significant acceleration from the target vehicle; ``Irregular left lane change" applies to scenarios in the highD dataset that involve a left lane change but do not fit the first two categories; ``Change to the right lane for overtaking" is when the target vehicle occupies the leftmost or middle lane and the lane ahead is blocked; ``Change right to the slow lane" is associated with significant deceleration from the target vehicle or when the target vehicle is a truck; ``Irregular right lane change" covers right lane change scenarios in the highD dataset that do not fit the first two rightward categories; ``Following and keep lane" occurs when following a slower vehicle ahead; ``Normal keep lane" is when the lane ahead is unobstructed and it is a lane keeping scenario in the highD dataset. These categories assist in predicting driving patterns and providing reliable explanations.

\begin{figure}
    \includegraphics [width=\linewidth, trim= 5 0 5 0, clip] {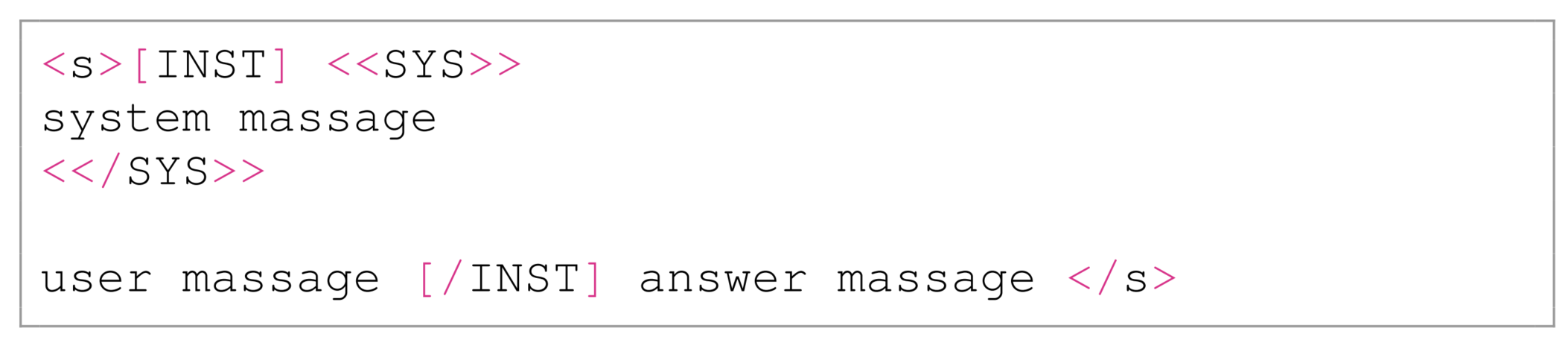}
    \caption{A data sample of Llama format. Special tokens, such as [INST], serve as delimiters distinguishing system messages, user messages, and answer messages. The details of these messages are elucidated in the Prompting section.}
    \label{fig:llama_data}
\end{figure}

\subsection{Fine-tuning}
In our work, we utilize an open-source foundational language model, Llama-2-13b-chat \cite{touvron2023llama}, as our pre-trained LLM. To achieve parameter-efficient finetuning, we adopt the Low-Rank Adaptation (LoRA) \cite{hu2021lora} strategy, which freezes the pre-trained model weights and injects trainable rank decomposition matrices into each layer of the Transformer architecture. Additionally, we customize the LLM for our specific prediction task using supervised fine-tuning techniques. The raw dataset utilized in our work originates from the highD dataset \cite{krajewski2018highd}, which captures human naturalistic vehicle trajectories on German highways. Following the fine-tuning instructions provided for Llama-2, each data sample is formatted to include an input prompt and a corresponding answer, separated by a special token. This formatting is illustrated in the Fig.~\ref{fig:llama_data}. The answer for each sample is derived from the ground truth obtained from the highD dataset, encompassing programmatically automatically labeled future trajectories, driving intentions and CoT reasoning. In summary, we process highD raw dataset with natural language, format each sample into the Llama format, and then feed it into LLM. Finally, we fine-tune the LLM by aligning the LLM's output $\{\mathcal{C}, \mathcal{I}, \mathcal{T}\}$ with the corresponding ground truth labels $\{\mathcal{C}^{*}, \mathcal{I}^{*}, \mathcal{T}^{*}\}$. This fine-tuning process is executed through the language modeling loss $\mathcal{L}$ as defined in Equation~\ref{equation:loss}. During fine-tuning, we mask the loss for tokens within the input prompts, focusing backpropagation only on the tokens comprising the answers, in alignment with the fine-tuning approach of Llama-2. By following this approach, our fine-tuned model, LC-LLM, can predict human driving behaviors and corresponding CoT reasoning. This capability is crucial for ensuring the safety of ego vehicles within an autonomous driving system.

\subsection{Interpretability of Prediction}
A prevalent limitation in contemporary autonomous driving prediction models lies in their constrained interpretability. This is attributed to the fact that these models generate predictions about the future behaviors or trajectories of the target vehicle through black-box neural networks, offering little explanation behind their prediction results. In our research, we address this limitation in two distinct ways. First, we fine-tune the CoT reasoning capabilities of our LC-LLM. By doing so, our model learns the domain knowledge and rules in driving, which enhances transparency and reliability of lane change predictions, leading to more accurate and explainable predictions. Second, we incorporate explanatory requirements into the input prompts in the inference stage, where the weights of our fine-tuned model are fixed. Benefiting from the self-explanation \cite{rajani2019explain, huang2023can} capabilities of LLMs, our fine-tuned model, LC-LLM, not only predicts lane change intentions and future trajectories but also provides corresponding CoT reasoning and explanations for its prediction results, thereby enhancing the interpretability of the prediction results. This approach facilitates a more transparent and understandable prediction process, which is crucial for the practical application of autonomous driving systems. Some visual examples of the interpretability of our LC-LLM model are shown in Fig.~\ref{fig:explanation}.

\section{Experimental Results and Analysis}
This section outlines the evaluations of our proposed LC-LLM model. First, we detail the dataset processing and the experimental setup. Subsequently, we elucidate the evaluation metrics. Next, we compare the proposed model with the baseline models and analyze the results quantitatively. Moreover, we perform an ablation study on the key components of our proposed method. Finally, we conduct an interpretability study and a zero-shot study to evaluate the interpretability and effectiveness of our approach.

\begin{table} 
\caption{Hyperparameters for training our LC-LLM.}
    \renewcommand\arraystretch{1.2}
		\begin{center}
			\begin{tabular}{m{4cm}|m{2.5cm}}
				\toprule
				\centering \textbf{Hyper-parameter} & \centering \textbf{Value} \cr
				\hline
                    \centering Learning rate& \centering 5e-4\cr
				\hline
                    \centering Batch size&\centering 8 \cr
                    \hline
                    \centering Train epoch&\centering 2 \cr
                    \hline
                    \centering Lora\_r& \centering 64 \cr
                    \hline
                    \centering Lora\_alpha& \centering 16 \cr
                    \hline
                    \centering Gradient\_accumulation\_steps& \centering 8 \cr
                    \hline
                    \centering Warmup\_steps& \centering 600 \cr
                    \hline
                    \centering Load\_in\_bits& \centering 8 \cr
                    \bottomrule
			\end{tabular}
			\label{tab:parameters}
		\end{center} 
\end{table}

\begin{table*}
\caption{Comparison of the intention prediction performance of the proposed LC-LLM model and baseline models on the highD dataset.} 
\begin{tabularx}{\textwidth}{p{1.7cm}|c|p{0.5cm}p{0.4cm}c|p{0.5cm}p{0.4cm}c|p{0.5cm}p{0.4cm}c|p{0.5cm}p{0.4cm}c|p{0.5cm}p{0.4cm}c}
\toprule 
\centering \multirow{2}{*}{Model} & \multirow{2}{*}{Intention} & \multicolumn{3}{c|}{$T \in[0, 1]$} & \multicolumn{3}{c|}{$T \in (1, 2 ]$} & \multicolumn{3}{c|}{$T \in (2, 3 ]$} &
\multicolumn{3}{c|}{$T \in (3, 4]$} & \multicolumn{3}{c}{Avg. ($T \in [0,4]$)} \\
    &   & P(\%)  & R(\%)   &  F1(\%)  & P(\%)   & R(\%)   &  F1(\%)  & P(\%)   & R(\%)   &  F1(\%)  & P(\%)   & R(\%)   &  F1(\%)  & P(\%)   & R(\%)   &  F1(\%) \\
\midrule
\centering \multirow{4}{*}{LSTM\cite{xin2018intention}}   & LK & 84.6 & 97.0  & 90.4  & 80.5  &  97.7 & 88.3  &  69.6  &  97.5 & 81.2  & 56.9  & 97.4  & 71.8  & 71.2  & 97.4 & 82.3 \\
                        & LLC & 97.8 & 83.0  & 89.8  & 98.6  &  79.3 & 87.9  &  97.1 &  67.3 & 79.5  & 95.3  & 49.9  & 65.5  &  97.4 & 69.9 & 81.4  \\
                        & RLC & 98.9 &  99.4 & 99.1  &  98.7 & 97.0  & 97.8  &  98.8 & 89.6  & 94.0  & 97.6  & 74.7  &  84.6 & 98.5  & 90.2  & 94.2  \\  
                        & Macro avg. & 93.8 & 93.1  &  93.1 &  92.6 & 91.3  &  91.3 &  88.5 & 84.8  & 84.9  & 83.3  & 74.0  & 74.0  & 89.0 & 85.8 & 85.9 \\
\hline
\centering \multirow{4}{*}{Transformer\cite{gao2023dual}}   & LK & 89.9 & 91.7  & 90.8  & 85.5  & 91.2  & 88.4  & 74.3  & 91.1  & 81.8  & 63.4  & 91.6  & 75.0  & 76.9  & 91.4 & 83.5 \\
                            & LLC & 95.6 & 91.1  & 93.3  &  94.6 & 85.7  & 89.9  &  93.5 & 73.8  & 82.4  & 92.7  & 62.4  & 74.6  & 94.2  & 78.2 & 85.5 \\
                        & RLC & 96.0 & 98.5  & 97.2  & 95.4  & 98.3  & 96.8  & 95.2  & 93.7  & 94.4  & 93.2  & 82.2  & 87.3  & 95.0  & 93.1 & 94.1 \\
                        & Macro avg. & 93.8 & 93.8  &  93.8 & 91.9  & 91.7  & 91.7  &  87.6 &  86.2 & 86.2  & 83.1  & 78.7  & 79.0  & 88.7  & 87.6 & 87.7 \\

\hline
\centering \multirow{4}{*}{LC-LLM}   & LK & 99.6 & 96.4  & 97.9  & 99.9  & 97.1  & 98.5  & 98.7  & 96.7  & 97.7  & 85.9  & 96.6  & 90.9  & 95.6  & 96.7 & 96.2 \\
                        & LLC & 97.9 & 99.4  & 98.6  & 98.6  & 99.6  & 99.1  & 97.7  & 98.7  & 98.2  &  97.1 & 89.3  & 93.0  & 97.8  & 96.7 & 97.3 \\
                        & RLC & 98.1 & 99.8  & 99.0  & 98.2  & 100.0  & 99.1  & 98.0  & 99.0  & 98.5  & 97.2  & 92.9  & 95.0  & 97.9  & 97.9 & 97.9 \\ 
                        & Macro avg. & 98.5 & 98.5 & \textbf{\underline{98.5}}  & 98.9  & 98.9  & \textbf{\underline{98.9}}  &  98.1 & 98.1 & \textbf{\underline{98.1}}  & 93.4  & 92.9  & \textbf{\underline{93.0}}  &  97.1 & 97.1 & \textbf{\underline{97.1}} \\
                        
\bottomrule
\end{tabularx}
\label{tab:result_comp}
\end{table*}

\subsection{Dataset Processing}
To evaluate the performance of the model proposed in this paper, the highD dataset was employed for both training and testing. The highD dataset is a large-scale natural vehicle trajectory dataset collected on German highways. It encompasses 16.5 hours of data from six distinct locations, involving 110,000 vehicles, covering a total distance of 45,000 km, and capturing 5,600 documented complete lane changes. The highD dataset comprises data extracted from 60 recordings. For our study, we select data from the first 50 recordings for training the model, while the remaining 10 recordings are reserved for testing. Furthermore, we adopt a target-centric strategy that normalizes all position inputs to the coordinate system centered on the current frame position of the target vehicle.

In our work, the lane change frame $\textit{T}_{lc}$ is defined as the frame in which the $lane\_id$ changes. The advanced prediction time $\textit{T}$ denotes the temporal interval between the lane change frame and the current frame, and it can be expressed as $\textit{T} = \textit{T}_{lc} - \textit{T}_{current}$, as illustrated in the Fig.~\ref{fig:advanced prediction time}. For the purpose of training and evaluating our proposed LC-LLM model, we extract both lane change (LC) and lane keeping (LK) scenarios from the highD dataset. The LK scenarios comprise data samples wherein the $lane\_id$ remains constant throughout, while the LC scenarios consist of data samples where the advanced prediction time $\textit{T}$ falls within the range of $[0,4]$ seconds. Furthermore, in order to evaluate the performance of our LC-LLM in different $\textit{T}$, we divide LC scenarios into four parts according to the four intervals of $\textit{T} (\textit{T}\in[0,1],\,\textit{T}\in(1,2],\,\textit{T}\in(2,3],\,\textit{T}\in(3,4])$. For the training dataset, we randomly select 48000 lane keeping (LK) samples, 12000 left lane change (LLC) samples from each $\textit{T}$ interval, and 12000 right lane change (RLC) samples from each $\textit{T}$ interval, resulting in a total of 144000 samples. Similarly, for the testing dataset, we randomly select 8000 LK samples, 2000 LLC samples from each $\textit{T}$ interval, and 2000 RLC samples from each $\textit{T}$ interval, accumulating a total of 24000 samples.

\begin{figure*}
    \centering
    \begin{minipage}{0.49\linewidth}
        \includegraphics[width=0.8\linewidth, trim=3 3 35 25, clip]{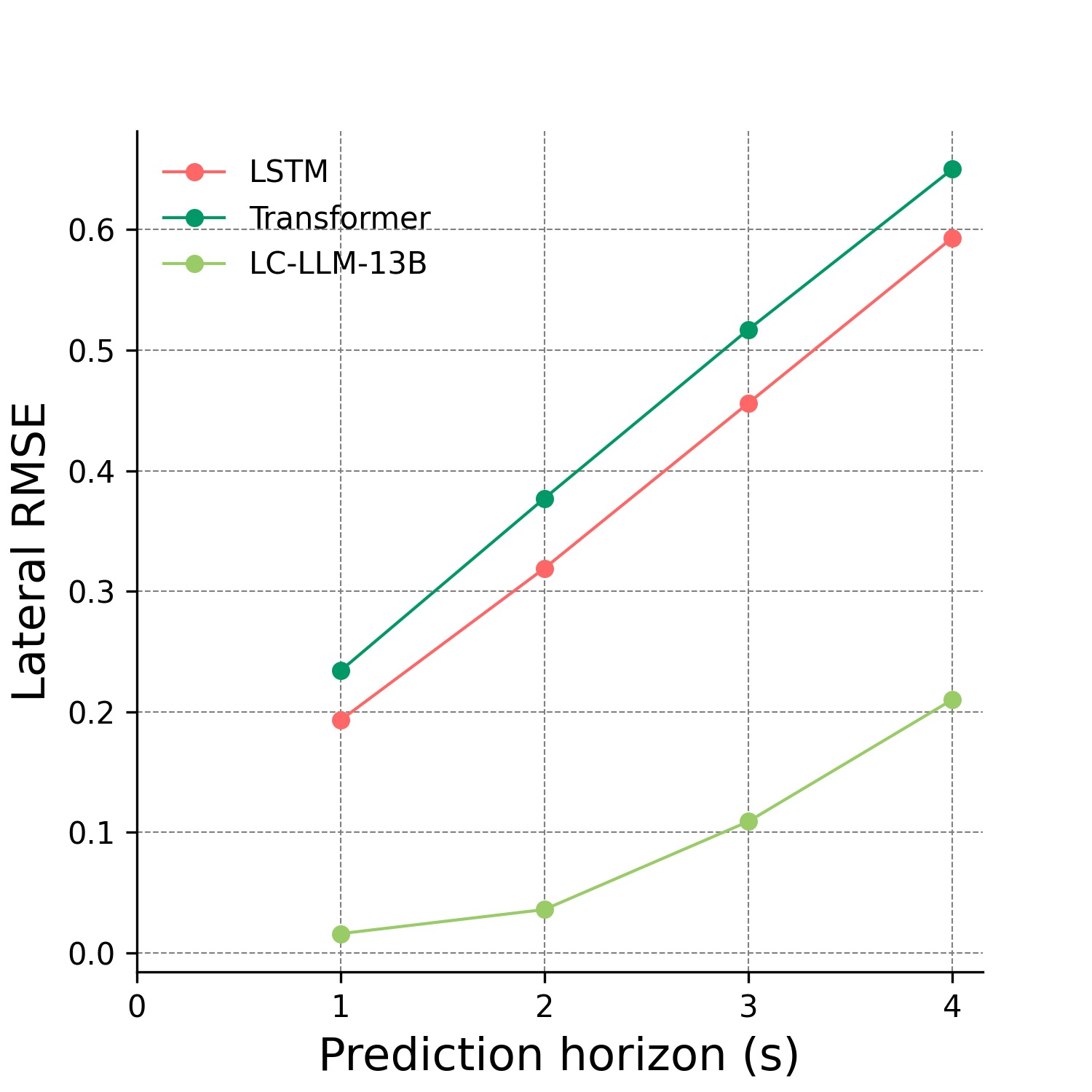}
        \caption*{(a) Lateral RMSE $\downarrow$}
        \label{fig:lateral_rmse}
    \end{minipage}
    \begin{minipage}{0.49\linewidth}
        \includegraphics[width=0.8\linewidth, trim=3 3 35 25, clip]{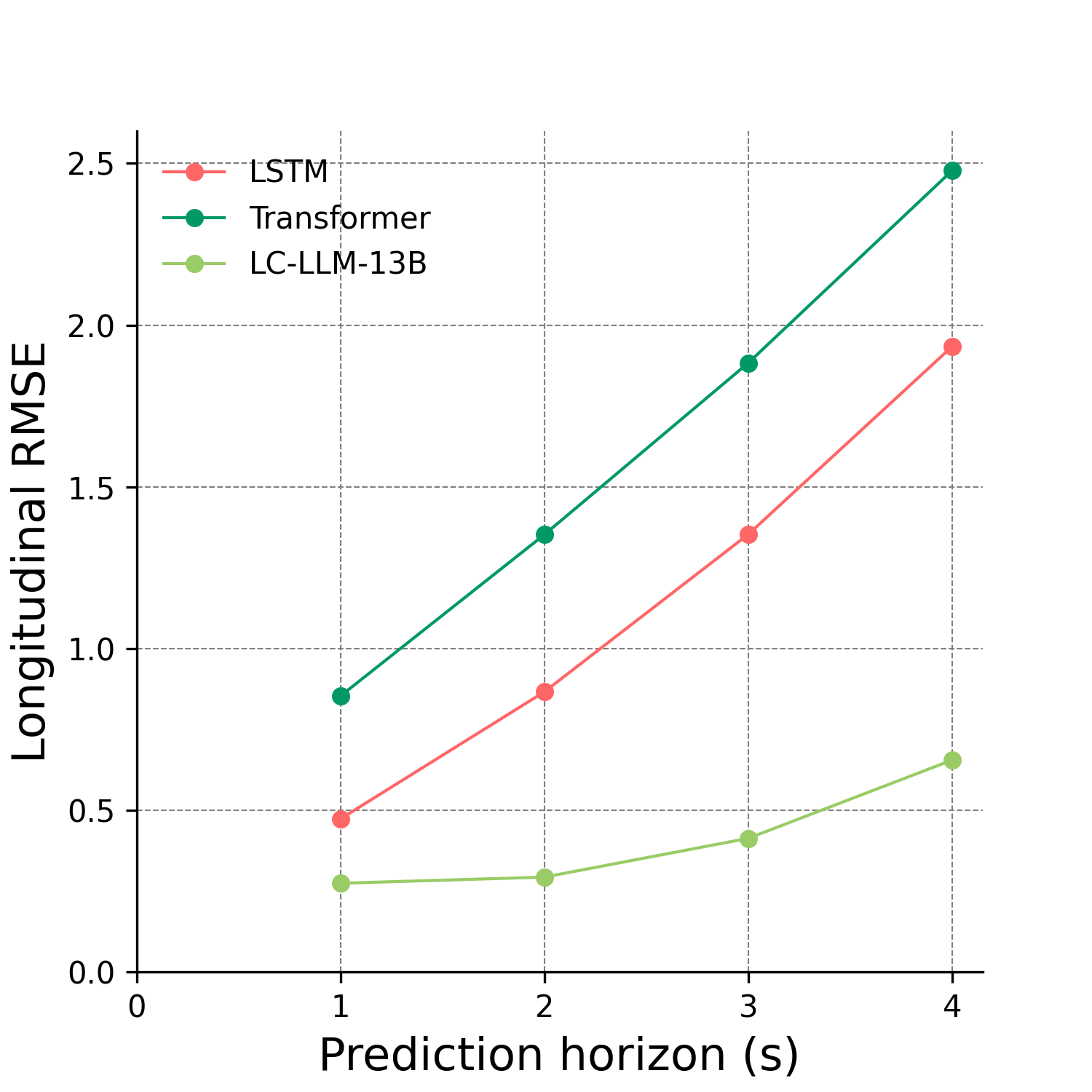}
        \caption*{(b) Longitudinal RMSE $\downarrow$}
        \label{fig:long_rmse}
    \end{minipage}
  \caption{Comparison of the trajectory prediction performance of the proposed LC-LLM model and baseline models on the highD dataset.}
  \label{fig:traj_result}
\end{figure*}

\subsection{Experimental Setup}
We utilize the training dataset extracted from the highD dataset to train our proposed LC-LLM model, as well as the reimplemented baseline models: an LSTM-based model\cite{xin2018intention} and a Transformer-based model\cite{gao2023dual}. All the models are implemented using the PyTorch framework\cite{paszke2019pytorch}. Additionally, we employ the DeepSpeed \cite{rasley2020deepspeed} library for distributed training of our LC-LLM model. The training process is conducted on eight A800 GPUs and takes 11 hours for the 13B model. The hyperparameters used for model training are detailed in Table~\ref{tab:parameters}. To ensure the stability of the experimental results, each experiment is repeated five times, and the average value is reported as the final result.

\subsection{Evaluation Metrics}

Similar to previous works, we utilize precision, recall, F1 score, and macro average metrics to evaluate the performance of the intention prediction task. Additionally, we use the root mean square error (RMSE) metric to evaluate the performance of the trajectory prediction task.

\subsubsection{Evaluation Metrics of Intention Prediction}

\begin{itemize}
    \item Precision: The ratio of correctly predicted positive instances to the total predicted positives. It is calculated as
    \[\text{Precision} = \frac{\text{True Positives}}{\text{True Positives + False Positives}}. \]

    \item Recall: The ratio of correctly predicted positive instances to the total actual positives. It is calculated as
    \[ \text{Recall} = \frac{\text{True Positives}}{\text{True Positives + False Negatives}}. \]

    \item F1 Score:  The harmonic mean of precision and recall. It is calculated as
    \[ \text{F1 Score} = 2 \times \frac{\text{Precision} \times \text{Recall}}{\text{Precision + Recall}}. \]

    \item Macro Avg: The average precision, recall, and F1 score calculated across all classes. It is calculated as the arithmetic mean of precision, recall, and F1 score.
\end{itemize}

\subsubsection{Evaluation Metrics of Trajectory Prediction}

\begin{itemize}
    \item RMSE (Root Mean Squared Error): A measure of the average deviation between predicted and actual values. It is calculated as
    \[ \text{RMSE} = \sqrt{\frac{\sum_{i=1}^{n}(y_i - \hat{y}_{i})^2}{n}}, \]
    where $y_i$ is the actual value, $\hat{y}_{i}$ is the predicted value, and $n$ is the number of instances.
\end{itemize}

In trajectory prediction evaluation, we use RMSE (lat) to assess lateral prediction error and RMSE (lon) to assess longitudinal prediction error.

\subsection{Overall Performance}

We evaluate our proposed LC-LLM model through two tasks, one is intention prediction and the other is trajectory prediction. The results are as follows.

\subsubsection{Intention Prediction Results and Analysis} \

Table~\ref{tab:result_comp} shows the overall performance of the proposed LC-LLM model as well as baseline models in intention prediction with four different advanced prediction time $\textit{T}$. We can note that our proposed LC-LLM significantly outperforms baseline models for all $\textit{T}$, especially for the longer $\textit{T}$. For example, when $\textit{T} \in (2, 3]$, the average F1 score of LC-LLM is 13.8\% higher than that of the Transformer model and 15.5\% higher than that of the LSTM. Additionally, when $\textit{T} \in (3,4]$, the average F1 score of LC-LLM surpasses that of the Transformer model by 17.7\% and outperforms the LSTM model by 25.7\%. These results demonstrate that our LC-LLM performs well even at the early stage of a lane change maneuver when lateral displacement is minimal.

\subsubsection{Trajectory Prediction Results and Analysis} \

We assessed the performance of our LC-LLM models in the trajectory prediction task and compared them with baseline models. The results are depicted in Fig.~\ref{fig:traj_result}. The left panel illustrates the trends of lateral RMSE for trajectory points across different models and prediction horizons, while the right panel shows the trends of longitudinal RMSE. The results indicate a notable improvement in both lateral and longitudinal RMSE for our LC-LLM models in comparison to the baselines. Notably, with larger prediction horizons, our model demonstrates increasing advantages over the baselines in lateral RMSE and longitudinal RMSE, indicating better robustness concerning long-term prediction horizons. 

The results of lane change intention prediction task and trajectory prediction task demonstrate that our LC-LLM outperforms all baseline models in long-term (4 s) prediction, achieving a 17.7\% improvement in lane change intention prediction, a 64.4\% improvement in lateral trajectory prediction, and a 66.1\% improvement in longitudinal trajectory prediction, respectively.

\begin{figure}[ht]
    \centering
    \includegraphics [width=\linewidth, trim=0 0 0 0, clip]{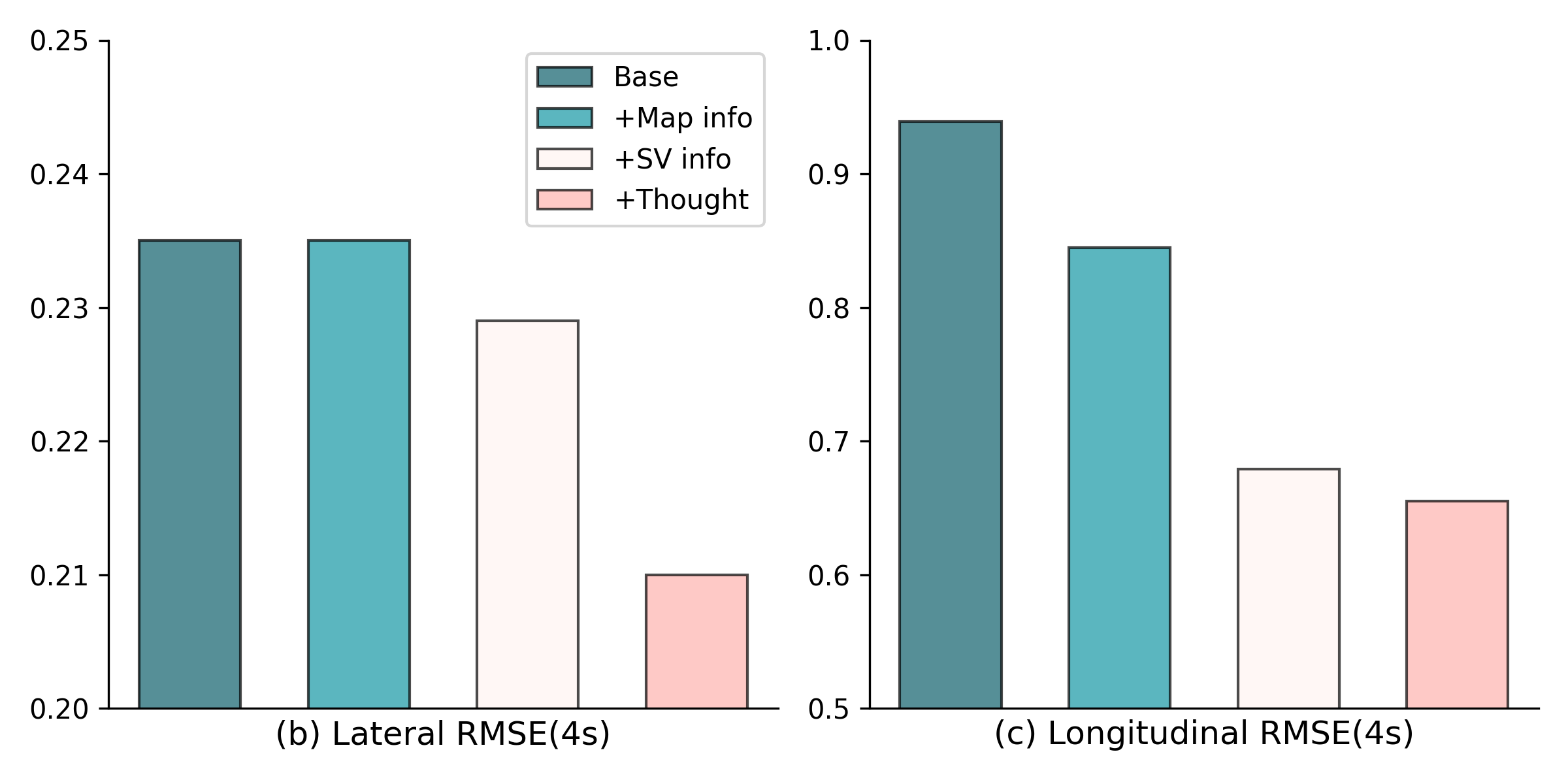}
    \caption{Components ablation studies. The ‘Base’ refers to the model results obtained using only the target vehicle information for prompting. The ‘+Map info’, ‘+SV info’, and ‘+Thought’ indicate the model results when map information, surrounding vehicle information, and CoT reasoning are sequentially incorporated, respectively.}
    \label{fig:prompt_ablation}
\end{figure}

\subsection{Ablation Study}

\subsubsection{Components Ablation Studies} \

We explored the impact of incorporating different components on the trajectory prediction task, and the results are illustrated in Fig.~\ref{fig:prompt_ablation}. A series of ablation experiments were conducted, progressively introducing different components—Target Vehicle Information, Map Information, Surrounding Vehicle Information, and Chain-of-Thought reasoning. The comprehensive model, integrating all contextual prompts and CoT reasoning component, exhibited superior performance compared to the base model with only Target Vehicle Information, achieving a substantial 10.64\% reduction in lateral RMSE, and a remarkable 29.57\% decrease in longitudinal RMSE. Specifically, the addition of Surrounding Vehicle Information proved crucial for the future trajectory forecasting. Furthermore, the incorporation of CoT reasoning yielded notable enhancements in trajectory prediction. 

\begin{table}[ht]
    \caption{Multi-task Ablation Study.} 
	\begin{center}
		\begin{tabular}{p{2.2cm}|p{1.8cm}|p{1.2cm}|p{1.2cm}} 
			\toprule 
			\centering \multirow{2}{*}{Task} & 
                \multicolumn{1}{c|}{Intention result} &\multicolumn{2}{c}{Trajectory result} \cr
			\centering & \centering F1 (avg)  & \centering RMSE(lat)  & \centering RMSE(lon)  \cr
			\midrule 
			\centering \multirow{1}{*}{Only Intention} 
			\centering & \centering 0.948& \centering --& \centering -- \cr
			\centering \multirow{1}{*}{Only Trajectory} 
			\centering & \centering --& \centering 0.257 & \centering 0.798 \cr
			\centering \multirow{1}{*}{Intention+Trajectory} 
			\centering & \centering \textbf{0.971}& \centering \textbf{0.210}& \centering \textbf{0.655} \cr
			\bottomrule 
		\end{tabular} 
	\end{center}
    \label{tab:multi_task}
\end{table}

\subsubsection{Multi-task Ablation Study} \

We further examined the efficacy of multi-task learning in concurrently predicting lane change intentions and future trajectories, as depicted in Table~\ref{tab:multi_task}. The findings demonstrate superior performance in the multi-task setting compared to the isolated trajectory prediction task, emphasizing the benefits of integrating high-level intention prediction for achieving more accurate trajectory prediction. Furthermore, in comparison to the standalone intention prediction task, adding trajectory prediction not only elevates the overall task complexity but also enables the model to capture more nuanced details, consequently, enhancing the accuracy of intention prediction tasks.

\begin{figure*}
\centering
\includegraphics[width=\linewidth, trim= 310 -10 165 0, clip]{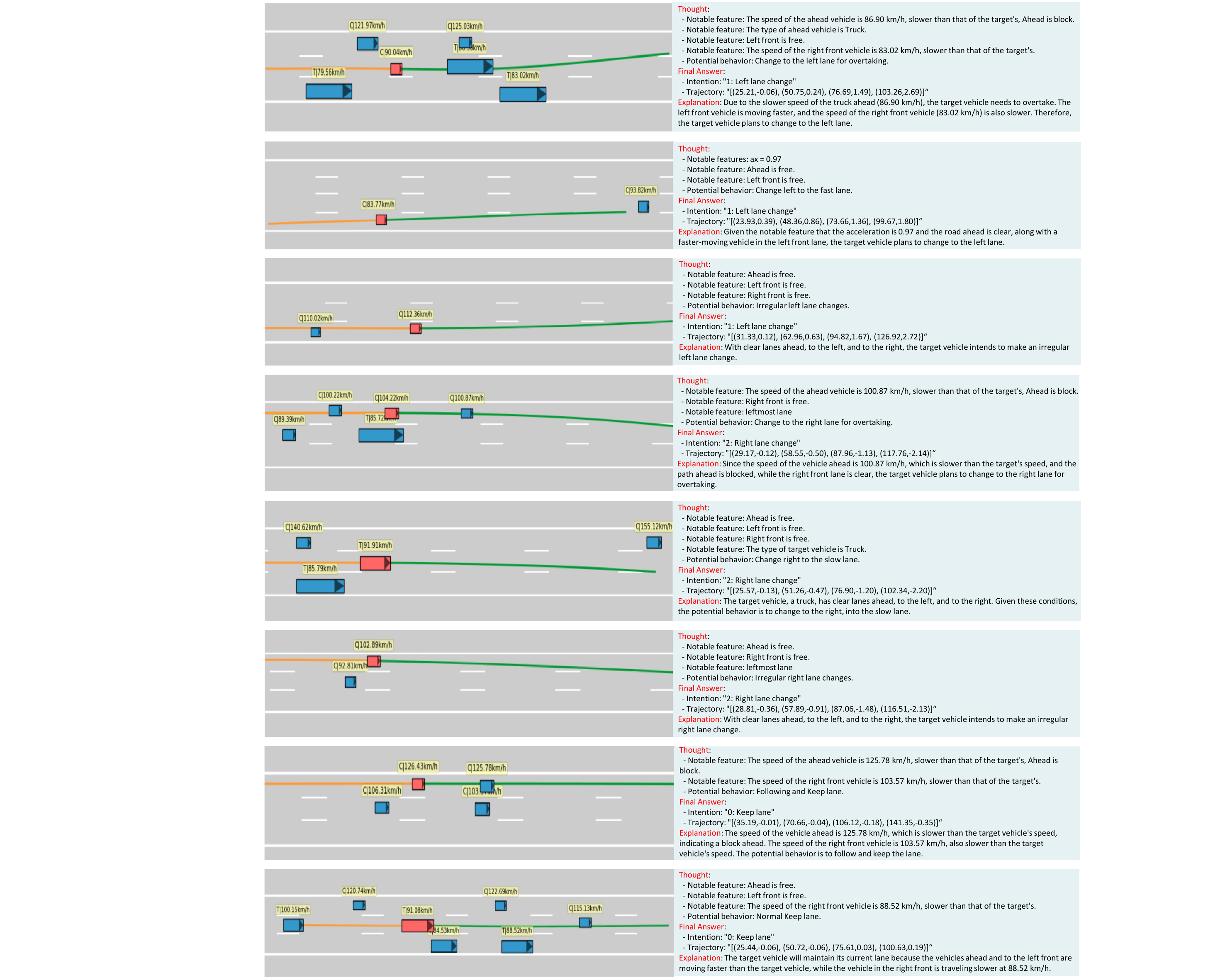}
\caption{Some visual examples demonstrating the interpretability of our LC-LLM model. In each example, the image on the left depicts the driving scenario, and the text on the right presents the corresponding output of our model, which includes the prediction results and their reasoning and explanations.}
\label{fig:explanation}
\end{figure*}

\subsection{Interpretability Study}
\subsubsection{Visualization of Results}
In order to demonstrate the interpretability of our LC-LLM, we visualize several sample scenarios alongside the corresponding outputs from our model, as depicted in Fig.~\ref{fig:explanation}. The visualization illustrates that our LC-LLM is capable of forecasting lane change intentions and future trajectories of the target vehicle while also providing corresponding CoT reasoning and explanations for the prediction results. For example, in the first case in the figure, with a slow-moving truck positioned ahead, the target vehicle need to overtaking to gain some speed advantage. And given the high velocity of the left front vehicle while the low velocity of right front vehicle, the target vehicle has an opportunity to execute a left lane change. The explanation given by our LC-LLM is consistent with this driving scenario. These explanations provided by our model for these scenarios serve as evidence that our LC-LLM possesses a deep understanding of the driving scenario and makes reasonable and reliable predictions for the target vehicle. In contrast to previous methodologies that solely generate prediction results, our LC-LLM model exhibit better interpretability.
\subsubsection{Quantitative Evaluation}
To quantitatively evaluate the interpretability of our LC-LLM, we evaluate the accuracy of CoT reasoning. We randomly selected 100 samples from the test dataset and manually compared the accuracy of the CoT reasoning output by our LC-LLM model with the ground truth in the dataset, scoring them accordingly. The scoring rule is that 10 points will be deducted for each error, omission, or addition of a notable feature, and 50 points will be deducted for a potential behavior error. Our quantitative evaluation result is that our model scores 97.2. This result demonstrates that our LC-LLM model excels in generating accurate and interpretable predictions, confirming the effectiveness of our methods in enhancing the transparency and reliability of autonomous driving systems.

\begin{table*}
    \centering
    \caption{Zero-shot Study.} 
    \renewcommand\arraystretch{1.2}
    \begin{tabular}{m{3.5cm}|m{2cm}|m{2cm}|m{3cm}|m{3cm}|m{2.5cm}}
    \toprule
        \centering \multirow{2}{*}{Method} & \multicolumn{2}{c|}{Intention Prediction} & \multicolumn{3}{c}{Trajectory Prediction} \cr
        & \centering F1 score(avg) & \centering Failed cases & \centering Lateral RMSE(4s) & \centering Longitudinal RMSE(4s) & \centering Failed cases \cr
        \hline
        \centering Zero-shot(Llama-13B-chat) & \centering 0.210 & \centering 4 & \centering 1.405 & \centering 63.366 & \centering 4  \cr
        \centering Fine-tune(LC-LLM) & \centering \textbf{0.971} & \centering \textbf{0} & \centering \textbf{0.210} & \centering \textbf{0.655} & \centering \textbf{0} \cr
    \bottomrule
    \end{tabular}
    \label{tab:zeroshot}
\end{table*}

\subsection{Zero-shot Study}
Table~\ref{tab:zeroshot} presents a comparative analysis between zero-shot and fine-tuned experiments across two tasks. The fine-tuned model demonstrated substantial improvements, yielding markedly higher F1 scores and significantly lower RMSE compared to the zero-shot experiment. These results underscore the efficacy of model fine-tuning in injecting domain-specific knowledge into Large Language Models. This affirms the suitability of fine-tuning as the preferred approach for enabling Large Language Models (LLMs) to excel in specific domains in future applications.

\subsection{Robustness Evaluations}
\begin{figure}
    \centering
    \includegraphics [width=\linewidth, trim=20 135 350 110, clip]{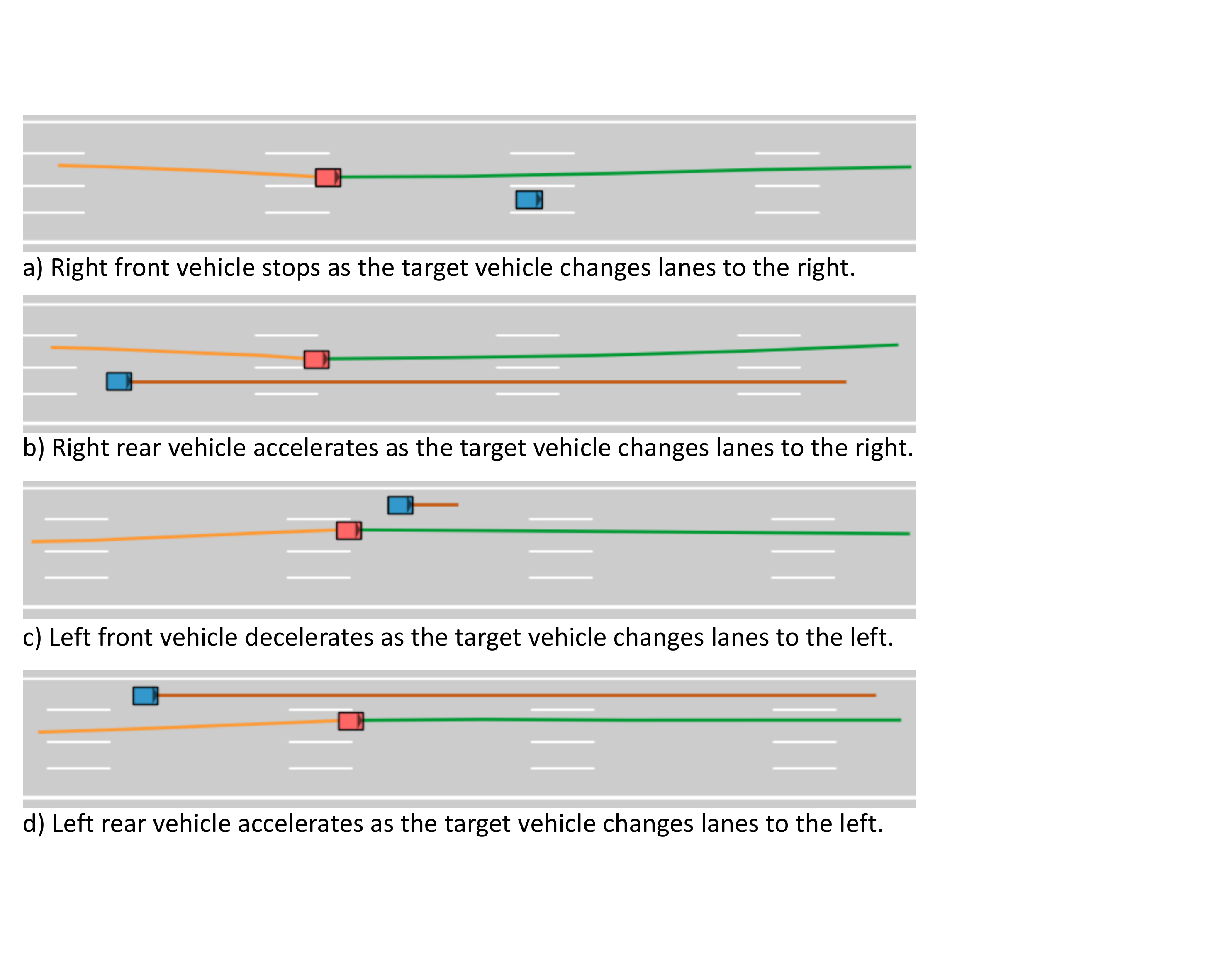}
    \caption{Some visual examples evaluating the robustness of our LC-LLM. These test samples are from outside the distribution of the highD dataset.}
    \label{fig:emergency}
\end{figure}

As mentioned in \cite{feng2023dense}, safety-critical events are important to test the model's robustness. In this experiment, we evaluated the robustness of our LC-LLM by creating four safety-critical scenarios not included in the HighD datasets. The first scenario involves a left lane change with the left front vehicle emergency braking, with speeds of 0 to 50 km/h in 10 km/h increments and relative distances of 10 to 100 meters in 10-meter increments, resulting in 60 samples. The second scenario tests a right lane change with the right front vehicle emergency braking, using the same speed and distance ranges, also resulting in 60 samples. The third scenario involves a left lane change with the left rear vehicle accelerating, with speeds of 100 to 150 km/h in 10 km/h increments and the same distance range, resulting in 60 samples. The fourth scenario examines a right lane change with the right rear vehicle accelerating, using the same speed and distance ranges as the third scenario, producing 60 samples.

The results, depicted in Fig.~\ref{fig:emergency}, highlight the robustness of our LC-LLM across these four challenging scenarios. In scenario (a), where the right front vehicle stops as the target vehicle changes lanes to the right, the LC-LLM accurately predicted the necessary adjustments to avoid a collision. Scenario (b) shows the right rear vehicle accelerating as the target vehicle changes lanes to the right; here, the LC-LLM successfully anticipated the increased speed of the rear vehicle and adjusted the trajectory accordingly. In scenario (c), the model handled the left front vehicle's deceleration effectively during the target vehicle's left lane change, maintaining safe distances and smooth transitions. Finally, in scenario (d), the LC-LLM demonstrated its ability to manage the left rear vehicle's acceleration during the target vehicle's left lane change, ensuring a safe maneuver. These results affirm the model's capability to handle out-of-distribution cases, enhancing its applicability in real-world autonomous driving scenarios.

\subsection{Limitations}

Our LC-LLM, while demonstrating significant improvements in lane change intention and trajectory prediction accuracy and interpretability, has limitations. Firstly, our model has been tested exclusively on the HighD dataset, which primarily comprises highway scenarios. To ensure the robustness and generalizability of our model, it is essential to evaluate its performance on more diverse datasets such as NuScenes \cite{nuscenes2019} and Waymo \cite{ettinger2021large}, which encompass a wider variety of complex urban and suburban traffic scenarios. While incorporating detailed map information from these complex urban scenarios into input prompts presents a considerable challenge. Consequently, the development of an appropriate prompt that accurately describe high-definition maps remains an objective for our future work. Secondly, our LC-LLM exhibits slower inference speed compared to baseline models. Specifically, the inference time for LC-LLM-13B is 26.72 seconds with a batch size of 32, significantly longer than the 0.011 seconds and 0.014 seconds for the LSTM and Transformer models, respectively. This slower inference time is partly due to the larger model parameters, posing a challenge for real-time applications. Future work should focus on optimizing the model for faster inference and exploring techniques such as knowledge distillation to reduce the computational load.

\section{Conclusion}
In this paper, we introduce LC-LLM, an explainable lane change prediction model that not only forecasts lane change intentions and trajectories but also provides CoT reasoning and explanations for its predictions. We reconceptualize the task of lane change prediction as a language modeling problem and employ a supervised fine-tuning technique to fine-tune the LLM specifically for this task. In this way, we successfully leverage the strong common sense reasoning capabilities and self-explanation abilities of the LLM to address the challenge of lane change prediction. Our extensive experiments conducted on highD dataset show that our LC-LLM improves the accuracy of predicting lane change intention and trajectory as well as significantly augments the interpretability of the prediction results. Future work includes extending our approach to urban driving scenarios, reducing inference times through knowledge distillation to improve model speed and efficiency, and developing methods to predict lane change intentions and trajectories for multiple vehicles simultaneously.

\section*{Acknowledgments}
This work is supported by the National Natural Science Foundation of China under Grant 52302379, Guangzhou Basic and Applied Basic Research Projects under Grants 2023A03J0106 and 2024A04J4290, Guangdong Province General Universities Youth Innovative Talents Project under Grant 2023KQNCX100, and Guangzhou Municipal Science and Technology Project 2023A03J0011.

\bibliographystyle{IEEEtran}
\bibliography{reference}

\vfill

\end{document}